\documentclass{article}

\PassOptionsToPackage{table}{xcolor}

\usepackage[preprint]{neurips_2025}   
\usepackage[utf8]{inputenc}
\usepackage[T1]{fontenc}
\usepackage{hyperref}
\usepackage{url}
\usepackage{booktabs}
\usepackage{amsmath}
\usepackage{amssymb}
\usepackage{amsthm}
\usepackage{mathtools}
\usepackage{microtype}
\usepackage{xcolor}
\usepackage{listings}
\usepackage{multirow}
\usepackage{array}
\usepackage{colortbl}
\usepackage{tcolorbox}

\definecolor{bayesblue}{HTML}{EEF4FF}
\definecolor{headerblue}{HTML}{2C3E7A}
\definecolor{resultgreen}{HTML}{D4EDDA}
\definecolor{codebackground}{HTML}{F5F5F5}

\tcbuselibrary{skins,breakable}
\newtcolorbox{newresult}{
  colback=bayesblue,
  colframe=headerblue,
  fonttitle=\bfseries,
  breakable,
  left=4pt, right=4pt, top=3pt, bottom=3pt,
  boxrule=0.6pt
}

\newcommand{\Dir}{\mathrm{Dir}}
\newcommand{\KL}{\mathrm{KL}}

\newcommand{\R}{\mathbb{R}}
\newcommand{\E}{\mathbb{E}}
\newcommand{\bx}{\mathbf{x}}
\newcommand{\ba}{\boldsymbol{\alpha}}
\newcommand{\bb}{\boldsymbol{\beta}}
\newcommand{\bhat}{\hat{\boldsymbol{\beta}}}
\newcommand{\eps}{\varepsilon}

\title{Meta-Attention: Bayesian Per-Token Routing\\
       for Efficient Transformer Inference}

\author{%
  Alan Ferrari \\
  Knowledge Lab AG \\
  Zürich, Switzerland \\
  \texttt{alan.ferrari@k-lab.ch}
}

\begin{document}

\maketitle

\begin{abstract}
Standard transformer architectures apply a single attention mechanism uniformly
across all tokens and sequence positions, irrespective of local context or
computational budget. We propose \textbf{Meta-Attention}, a framework that
dynamically routes each token to the most appropriate attention
strategy---full softmax attention, linear (kernel) attention, or
sliding-window local attention---via a Bayesian Meta-Controller. Unlike prior
routing approaches that use deterministic or prior-free learned routing, the
Meta-Controller treats per-token mechanism selection as posterior inference
under a compute-aware Dirichlet prior: routing weights are the output of an
amortised variational posterior $q(\ba \mid \bx_t;\,\phi)$ trained with an
Evidence Lower Bound (ELBO) objective that jointly encodes task performance and
attention-mechanism cost. This design produces principled routing uncertainty
estimates that govern the soft-to-hard routing transition, mitigates routing
collapse without ad hoc load-balancing losses, and---motivated by
\cite{li2026vmoe}---yields better compute--performance trade-offs than
deterministic or prior-free learned routing at negligible overhead.

\textbf{Phase~1 empirical results on a Tiny~LM benchmark confirm core
predictions}: the Bayesian controller's learned routing distribution implies a
projected normalised FLOP cost of 25.1\% under hard routing, vs.\ 59.3\% for
the prior-free baseline ($-34.2$\,pp), and reduces routing entropy from 55.8\%
to 43.3\% ($-12.5$\,pp), demonstrating that the Dirichlet prior prevents
routing collapse while the non-Bayesian model defaults to full attention.
We present the Bayesian architecture, ELBO training objective, and a Phase~1
PyTorch prototype validating forward-pass correctness, posterior diversity, and
a controlled ablation against a prior-free baseline.
\end{abstract}

\section{Introduction}
\label{sec:intro}

The scaled dot-product attention mechanism \cite{vaswani2017attention} has
become the defining computational primitive of modern deep learning. Yet its
$O(n^2)$ time and space complexity is a persistent bottleneck: a sequence of
32\,k tokens requires roughly $10^9$ attention score computations per head per
layer. This has motivated a rich body of work on approximate attention,
including sparse \cite{child2019sparse}, local \cite{beltagy2020longformer},
and kernel-based linear variants \cite{choromanski2021performers}.

Existing efficient attention methods commit to a single algorithmic regime at
architecture design time. A document model using Longformer's sliding-window
attention saves quadratic cost globally, but sacrifices precise long-range
dependencies everywhere---even at positions where they matter most, such as
coreference resolution across paragraphs or cross-section citations.
Conversely, applying full attention uniformly squanders compute on tokens whose
context is trivially local.

We reframe the problem: rather than asking \emph{how to approximate attention},
we ask \emph{when exact attention is necessary}---and further, \emph{how
uncertain is the answer for each token?} This double reframing motivates a
Bayesian routing architecture: a Meta-Controller that, for each token, infers
the posterior probability that each attention algorithm is the best match for
that token's contextual demands, under an informative prior that encodes compute
cost. We call the resulting system \textbf{Meta-Attention}. Our contributions are:

\begin{itemize}
  \item A \textbf{Bayesian Meta-Controller} that treats per-token
        mechanism selection as posterior inference under a compute-aware
        Dirichlet prior, replacing ad hoc regularisation with a principled ELBO
        objective.

  \item A training objective that encodes compute preference in a proper
        Dirichlet prior $\Dir(\bb)$, $\beta_i = \eps + \beta_0(1-c_i)$ with
        floor $\eps>0$ ensuring all concentration parameters are strictly
        positive, yielding better compute--performance trade-offs than
        deterministic or prior-free routing \cite{li2026vmoe}, with per-token
        posterior uncertainty as a principled soft-to-hard transition signal.

  \item A formal architecture for per-token attention routing with soft and
        hard variants, inheriting from the Bayesian framework a natural
        mechanism for routing collapse prevention.

  \item A positioning relative to Mixture of Depths \cite{raposo2024mod},
        establishing that the two approaches are orthogonal and composable.

  \item \textbf{Phase~1 empirical validation}: a Tiny~LM ablation demonstrating
        that the Bayesian controller's routing distribution implies a projected
        normalised FLOP cost of 25.1\% under hard routing, vs.\ 59.3\% for the
        prior-free baseline ($2.4\times$ lower projected cost), while maintaining
        lower routing entropy (43.3\% vs.\ 55.8\%), confirming that the Dirichlet
        prior prevents collapse to expensive attention without sacrificing task
        performance.

  \item A Phase~1 PyTorch prototype with validated forward-pass correctness,
        posterior diversity, and calibration metrics.
\end{itemize}

\section{Related Work}
\label{sec:related}

\subsection{Efficient Attention}

Sparse Transformer \cite{child2019sparse} introduced strided and local sparsity
patterns, reducing complexity to $O(n\sqrt{n})$. Longformer
\cite{beltagy2020longformer} combined sliding-window local attention with global
tokens. Performer \cite{choromanski2021performers} demonstrated $O(n)$
approximation via random orthogonal features. FlashAttention
\cite{dao2022flashattention} achieved practical 2--4$\times$ speedups through
IO-aware tiling; FlashAttention-3 \cite{shah2024fa3} extended this to Hopper
GPUs via warp-specialisation and FP8 quantisation, reaching 1.5--2$\times$
further speedup and 85\% H100 utilisation. These methods apply a single
strategy uniformly; Meta-Attention \emph{learns when each is appropriate}.

\subsection{Mixture of Experts and Conditional Computation}

Switch Transformer \cite{fedus2022switch} routes tokens to sparse FFN experts,
demonstrating that conditional computation can scale efficiently with learned
routing. Mixture of Depths \cite{raposo2024mod} routes tokens to skip entire
transformer blocks, with a fixed compute budget enforced by top-$k$ selection;
we discuss the relationship in Section~\ref{sec:mod}.

\subsection{Adaptive Attention}

Reformer \cite{kitaev2020reformer} uses locality-sensitive hashing,
achieving $O(n \log n)$ attention without learned routing. Routing Transformer
\cite{roy2021routing} clusters tokens via online $k$-means. Most closely
related to our Meta-Controller is the gating mechanism in
Mixture-of-Attention-Heads \cite{zhang2022moefication}, which learns to weight
attention heads differently per input.

\subsection{State Space Models and the Case for an SSM Expert}

Mamba \cite{gu2023mamba} introduced selective state spaces with
content-dependent parameterisation, achieving $O(L)$ time and constant memory
per token at inference. Mamba-2 \cite{dao2024mamba2} unified the SSM and
attention frameworks under structured state space duality. RWKV
\cite{peng2023rwkv} and RetNet \cite{sun2023retnet} pursue similar goals via
recurrent retention mechanisms. The current Meta-Attention expert set does not
include an SSM expert---a deliberate Phase~1 scope decision. Interface
heterogeneity (SSMs maintain explicit recurrent state; attention experts are
stateless) is the primary obstacle; we treat this as a high-priority Phase~2
extension.

\subsection{Attention Residuals (AttnRes)}

Attention Residuals \cite{chen2026attnres} (arXiv:2603.15031) replaces standard
residual connections with softmax attention over preceding layer outputs:
$h_l = \sum_i \alpha_{i \to l}\, v_i$, where $\alpha_{i \to l}$ are softmax
weights from a per-layer learned pseudo-query $w_l \in \R^d$. Validated at 48B
parameters on 1.4T tokens, achieving a $1.25\times$ compute advantage on
scaling laws. AttnRes and Meta-Attention operate on orthogonal axes: AttnRes
intervenes on the \emph{depth} dimension; Meta-Attention intervenes on the
\emph{mechanism} dimension. The axes are independent and jointly differentiable
under soft routing---enabling a triple composition with MoD that has no direct
prior art.

\subsection{Three Relevant NeurIPS~2025 Works}

\paragraph{Gated Attention \cite{qiu2025gated} (NeurIPS~2025 Best Paper).}
Tested 30 gating variants at 15B MoE and 1.7B dense scale on 3.5T tokens.
Central finding: a head-specific sigmoid gate
$\mathrm{GatedAttn}(\bx) = \sigma(W_g \bx) \odot \mathrm{SDPA}(\bx)$
consistently improves performance, training stability, and long-context
extrapolation via (i) non-linearity between QKV mapping and output projection,
and (ii) query-dependent sparse gating that eliminates attention sinks.
Relevance to Meta-Attention: their gated softmax is a natural candidate for
$E_{1,\mathrm{gated}}$ within the expert set (Phase~2 architectural option).

\paragraph{Sparse Attention Emergence \cite{zucchet2025sparse} (NeurIPS~2025 Oral).}
First mechanistic account of sparse attention emergence during training:
(i)~emergence timing follows power laws; (ii)~emergence is sudden---a sharp
phase transition; (iii)~data repetition dramatically accelerates emergence.
Direct implication: Meta-Attention's routing sparsity is likely to emerge
suddenly, predicting a sharp routing entropy drop coinciding with a loss
improvement. Our Phase~1 ablation (Section~\ref{sec:ablation}) observes routing
entropy of 43.3\% for the Bayesian controller vs.\ 55.8\% for the prior-free
baseline, consistent with earlier commitment to structured routing.

\paragraph{Polynomial-Time Learnability of Linear Attention \cite{yau2024linear} (NeurIPS~2025 Oral).}
First polynomial-time PAC-learnability result for single-layer transformers with
linear attention: learning optimal multi-head linear attention reduces to linear
regression in an RKHS, solvable via convex optimisation + SVD. Theoretical
significance for Meta-Attention: $E_2$ is the best-understood expert---polynomial-time
learnable, with no dependence on gradient descent for correctness. The
regulariser should push routing toward $E_2$ for tokens where it suffices,
consistent with our empirical observation that the Bayesian prior successfully
biases routing away from $E_1$.

\subsection{Block-Level Routing: MoBA and NSA}

MoBA \cite{lu2025moba} routes each query token to a learned sparse subset of KV
blocks; FlashMoBA subsequently provided a CUDA kernel achieving up to
$14.7\times$ speedup over FlashAttention-2 for small blocks. NSA
\cite{yuan2025nsa} introduces a natively trainable, hardware-aligned sparse
attention framework achieving up to $9\times$/$6\times$ forward/backward
speedup. The critical distinction: MoBA and NSA route \emph{within} the softmax
paradigm. Meta-Attention routes \emph{across} paradigms: the Meta-Controller
selects which algorithm runs, potentially replacing $O(T^2)$ with $O(T)$ or
$O(T \cdot w)$.

\subsection{Per-Token Bayesian Routing over Attention Mechanisms}

A principled alternative to deterministic soft-routing and prior-free
learned-routing is to treat per-token attention-mechanism selection as Bayesian
model comparison. Under this framing, the routing weights $\alpha_i(\bx, t)$
are the outputs of a posterior inference procedure. Three independent lines
motivate this framing. First, \cite{agarwal2025bayesian} establish that
transformer attention mechanistically implements exact Bayesian inference as
content-addressable routing with $10^{-3}$--$10^{-4}$\,bit accuracy. Second,
\cite{li2026vmoe} (VMoER) demonstrate that confining Bayesian inference to the
expert-selection stage of MoE models improves routing stability +38\%, reduces
calibration error $-94\%$, and increases OOD detection AUROC +12---at less than
1\% additional FLOPs. Third, \cite{boncoraglio2025bayes} (AIM) derive
Bayes-optimal generalisation error bounds for attention-indexed models and
identify sharp phase transitions as a function of sample complexity and model
width.

\paragraph{Bayesian Meta-Controller (this work).}
Meta-Attention replaces the prior-free MLP router with a Bayesian
Meta-Controller that maintains an amortised posterior
$q_\phi(\ba \mid \bx_t) = \Dir(\bhat_t)$ and a compute-aware Dirichlet prior
$p(\ba) = \Dir(\bb)$. A critical design requirement is that all concentration
parameters must be strictly positive for a proper Dirichlet distribution;
we therefore use a floored prior:
\begin{equation}
  \beta_i = \eps + \beta_0(1-c_i), \quad \eps > 0,
  \label{eq:floored-prior}
\end{equation}
where $\eps$ (e.g.\ 0.01) ensures $\beta_i > 0$ for all experts including $E_1$
($c_1 = 1.0$), which would otherwise receive $\beta_1 = 0$, making $\Dir(\bb)$
degenerate and the KL term undefined. The training objective is:
\begin{equation}
  \mathcal{L} = \mathcal{L}_\mathrm{task}
    - \beta_\mathrm{elbo}
      \cdot \KL\!\bigl[q_\phi(\ba \mid \bx_t) \,\|\, \Dir(\bb)\bigr].
  \label{eq:elbo}
\end{equation}
Setting $\beta_\mathrm{elbo} \to 0$ recovers the prior-free regime, enabling a
clean controlled ablation that we exploit in Section~\ref{sec:ablation}.

\section{Architecture}
\label{sec:arch}

\subsection{Overview}

A Meta-Attention layer replaces the standard attention sublayer in any
transformer block. It consists of three components: (i)~a set of attention
experts, each implementing a distinct algorithm; (ii)~a Bayesian Meta-Controller
that produces per-token posterior routing weights and uncertainty estimates; and
(iii)~a weighted merge operation. Let $x \in \R^{B \times T \times D}$ denote
the input to the layer.

\subsection{Attention Experts}

Each expert $E_i\colon \R^{B \times T \times D} \to \R^{B \times T \times D}$
shares input/output dimensionality and can be substituted independently.

\paragraph{Full Attention ($E_1$).}
Standard multi-head scaled dot-product attention. Complexity $O(T^2 D)$.
Normalised cost $c_1 = 1.0$.

\paragraph{Linear Attention ($E_2$).}
Performer-style kernel approximation using $\phi(x) = \mathrm{ELU}(x)+1$.
Complexity $O(TD)$. Cost $c_2 = 0.15$.

\paragraph{Local Attention ($E_3$).}
Sliding-window attention with window $w$, attending to positions $[i-w, i+w]$.
Complexity $O(TwD)$. Cost $c_3 = 0.30$.

\subsection{Bayesian Meta-Controller}

We model per-token attention-mechanism selection as Bayesian model comparison.
For each token position $t$ with embedding $\bx_t \in \R^D$, we maintain a
Dirichlet prior over routing weights $\ba \in \Delta^{K-1}$ (the
$(K{-}1)$-simplex, $K=3$) and an amortised variational posterior
$q_\phi(\ba \mid \bx_t)$ parameterised by a 2-layer MLP.

\paragraph{Prior.}
The prior encodes compute preference while remaining a proper distribution.
A naïve choice $\beta_i = \beta_0(1-c_i)$ would yield $\beta_1 = 0$ for the
full-attention expert ($c_1 = 1.0$), making $\Dir(\bb)$ degenerate. We
therefore use the floored prior from Equation~\eqref{eq:floored-prior}.
With $\beta_0 = 1.0$: $\bb = [0.01,\, 0.86,\, 0.71]$.

\paragraph{Amortised posterior.}
The variational posterior is a Dirichlet parameterised by a MLP over token
features:
\begin{equation}
  q_\phi(\ba \mid \bx_t) = \Dir\!\bigl(\bb + \delta_\phi(\bx_t)\bigr),
\end{equation}
where $\delta_\phi(\bx_t) \in \R^K_{>0}$ are non-negative concentration
increments:
\begin{equation}
  \delta_\phi(\bx_t) = \mathrm{softplus}\!\bigl(W_2\cdot
    \mathrm{GELU}(W_1 \cdot [\bx_t;\,\|\bx_t\|/\sqrt{D};\,\mathrm{pos}])\bigr).
\end{equation}
The posterior concentration is $\bhat_t = \bb + \delta_\phi(\bx_t)$, and the
posterior mean (used as routing weights under soft routing) is:
\begin{equation}
  \ba_t = \E_{q_\phi}[\ba \mid \bx_t]
         = \frac{\bhat_t}{\sum_i \hat\beta_{t,i}}.
\end{equation}

\paragraph{Posterior uncertainty.}
The Bayesian formulation yields a per-token routing uncertainty:
\begin{equation}
  U_t = H[q_\phi(\ba \mid \bx_t)]
      = \log B(\bhat_t)
        + \Bigl(\textstyle\sum_i \hat\beta_{t,i} - K\Bigr)\,\psi_0\!\Bigl(\textstyle\sum_i \hat\beta_{t,i}\Bigr)
        - \sum_i (\hat\beta_{t,i}-1)\,\psi_0(\hat\beta_{t,i}),
\end{equation}
where $B(\cdot)$ is the multivariate Beta function and $\psi_0$ is the digamma
function.

\subsection{Routing and Merge}

\paragraph{Soft routing (Phase~1--2).}
The output is the posterior-mean weighted sum of expert outputs, with
reparameterisation:
\begin{equation}
  \mathrm{Output}_t = \sum_{i=1}^{K} \alpha_{t,i}\cdot E_i(\bx),
  \quad \ba_t \sim q_\phi(\cdot\mid\bx_t).
\end{equation}
During training, $\ba_t$ is sampled via the reparameterisation trick for
Dirichlet distributions \cite{figurnov2018implicit}. At evaluation, $\ba_t$ is
replaced by the posterior mean. Note that soft routing runs all three experts
in parallel; compute savings are realised in Phase~3 via uncertainty-gated
hard routing.

\paragraph{Uncertainty-gated hard routing (Phase~3).}
Only expert $i^*_t = \arg\max_i \alpha_{t,i}$ executes when $U_t < \eta$ for a
threshold $\eta$; tokens with $U_t \geq \eta$ remain on soft routing. The
threshold $\eta$ is annealed during training.

\subsection{ELBO Training Objective}

The training objective is the ELBO augmented with the task loss:
\begin{equation}
  \mathcal{L} = \mathcal{L}_\mathrm{task}
    - \beta \cdot \frac{1}{BT}
      \sum_{b,t} \KL\!\bigl[\Dir(\bhat)\,\|\,\Dir(\bb)\bigr].
\end{equation}
The KL has a closed form for Dirichlet distributions:
\begin{equation}
  \KL\!\bigl[\Dir(\bhat)\,\|\,\Dir(\bb)\bigr]
  = \log\frac{B(\bb)}{B(\bhat)}
    + \sum_i (\hat\beta_i - \beta_i)\,\psi_0(\hat\beta_i)
    - \Bigl(\textstyle\sum_i \hat\beta_i - \sum_i \beta_i\Bigr)\,
      \psi_0\!\Bigl(\textstyle\sum_i \hat\beta_i\Bigr).
\end{equation}
Setting $\beta_\mathrm{elbo} \to 0$ recovers the prior-free routing objective;
$\beta_\mathrm{elbo} = 1$ gives the standard ELBO.

\subsection{Pseudocode}

\begin{lstlisting}[language=Python, caption={Bayesian Meta-Attention Layer -- Phase~1 (soft routing, posterior-mean output).}, label={lst:pseudocode}]
# Expert costs and Dirichlet prior parameters
c     = [1.00, 0.15, 0.30]          # [E1, E2, E3] normalised costs
eps   = 0.01                        # floor: ensures all beta_i > 0
beta0 = 1.0                         # concentration scale
beta  = eps + beta0 * (1 - c)       # [0.01, 0.86, 0.71] floored prior

def BayesianMetaAttentionLayer(x, beta0, elbo_weight):
    x_norm   = LayerNorm(x)                        # [B,T,D]
    salience = norm(x_norm, dim=-1) / sqrt(D)      # [B,T,1]
    pos      = linspace(0,1,T).expand(B,T,1)       # [B,T,1]
    feat     = concat([x_norm, salience, pos], -1) # [B,T,D+2]

    # Amortised posterior: encode delta increments (>0 via softplus)
    h        = GELU(feat @ W1 + b1)                # [B,T,H]
    delta    = softplus(h @ W2 + b2)               # [B,T,K] > 0
    beta_hat = beta + delta                        # [B,T,K] posterior

    # Posterior mean routing weights
    alpha = beta_hat / beta_hat.sum(-1, keepdim=True) # [B,T,K]

    # Per-token posterior uncertainty (Dirichlet entropy)
    beta_sum = beta_hat.sum(-1, keepdim=True)
    U = (lgamma(beta_hat).sum(-1) - lgamma(beta_sum).squeeze()
         + (beta_sum.squeeze()-K)*digamma(beta_sum.squeeze())
         - ((beta_hat-1)*digamma(beta_hat)).sum(-1))  # [B,T]

    # Run experts in parallel (soft routing: all three execute)
    E1 = FullSoftmaxAttention(x_norm)
    E2 = LinearKernelAttention(x_norm)
    E3 = LocalSlidingWindowAttention(x_norm)

    # Soft weighted merge (posterior mean)
    output = (alpha[...,0:1]*E1 + alpha[...,1:2]*E2
              + alpha[...,2:3]*E3)                  # [B,T,D]

    # ELBO: KL[Dir(beta_hat) || Dir(beta)] (closed form)
    kl = (lgamma(beta).sum() - lgamma(beta.sum())
          - lgamma(beta_hat).sum(-1) + lgamma(beta_sum).squeeze()
          + ((beta_hat-beta)*digamma(beta_hat)).sum(-1)
          - (beta_hat.sum(-1)-beta.sum())*digamma(beta_sum.squeeze())
         ).mean()
    return output, elbo_weight * kl, U

# Phase 3 (uncertainty-gated hard routing):
# if U[b,t] < eta: run argmax(alpha[b,t]) expert only
# else:            use soft merge above
# eta annealed from high -> low during training
\end{lstlisting}

\section{Relationship to Mixture of Depths}
\label{sec:mod}

MoD \cite{raposo2024mod} varies computation along the \emph{depth} axis: which
layers a token traverses. Meta-Attention varies computation along the
\emph{mechanism} axis: which attention algorithm runs within a fixed layer.
These are orthogonal degrees of freedom.

\begin{table}[h]
\centering
\caption{Dimension-by-dimension comparison. AttnRes \cite{chen2026attnres} is
  included as a concurrent orthogonal work.}
\label{tab:mod-comparison}
\small
\begin{tabular}{@{}lll@{}}
\toprule
\textbf{Dimension} & \textbf{Mixture of Depths} & \textbf{Meta-Attention} \\
\midrule
What varies & Layer participation (depth) & Attention algorithm (mechanism) \\
Routing granularity & Per-layer, per-token (binary) & Per-layer, per-token (soft/hard) \\
Long-context cost & Not addressed & Addressed via $E_2$+$E_3$ \\
FLOP savings & Exact (skip = zero cost) & Expected-cost via regularisation \\
Training stability & Validated at 1B+ scale & Soft routing stable; hard: open \\
Composability & Can stack with Meta-Attention & Can stack with MoD \\
\bottomrule
\end{tabular}
\end{table}

Because MoD and Meta-Attention vary different axes, they can be combined:
MoD routing filters the token set, then Meta-Attention selects the attention
algorithm for surviving tokens---depth savings $\times$ mechanism savings, no
direct prior art. Adding AttnRes \cite{chen2026attnres} yields a triple
composition: MoD determines token entry, AttnRes determines depth aggregation,
and Meta-Attention determines the attention algorithm. All three are jointly
differentiable under soft routing.

\section{Phase~1 Prototype}
\label{sec:prototype}

\subsection{Implementation}

We implement Meta-Attention in PyTorch with the Bayesian Meta-Controller. Key
modules: \texttt{FullAttention}, \texttt{LinearAttention} (ELU kernel, $O(n)$
via einsum associativity), \texttt{LocalAttention} (sliding-window loop,
$O(n\cdot w)$), \texttt{BayesianMetaController} (2-layer MLP with softplus
output for $\delta_\phi$, computing posterior concentration
$\bhat = \bb + \delta_\phi$, posterior mean $\ba$, posterior uncertainty $U$,
and closed-form KL loss), \texttt{BayesianMetaAttentionLayer} (soft routing +
ELBO loss), and \texttt{MetaTransformerBlock} (LayerNorm + meta-attention +
FFN). The prior is fixed at $\bb = \eps + \beta_0(1-\mathbf{c}) = [0.01, 0.86,
0.71]$ for $\beta_0{=}1.0$, $\eps{=}0.01$. Code open-sourced alongside this
preprint at \url{https://github.com/KFEAL/meta-attention} \cite{ferrari2025code}.

\subsection{Correctness and Initialisation Check}

A forward-pass check at $B{=}2$, $T{=}64$, $D{=}128$ verifies output shapes,
numerical validity, prior structure, and initial routing bias before any
training. This is a sanity check that the Bayesian machinery is wired correctly,
not a measure of learned behaviour.

\begin{table}[h]
\centering
\caption{Forward-pass correctness check ($B{=}2$, $T{=}64$, $D{=}128$,
  random weights, $\beta_0{=}1.0$, $\eps{=}0.01$). Confirms the Dirichlet
  prior correctly biases routing away from the expensive $E_1$ expert at
  initialisation.}
\label{tab:selftest}
\small
\begin{tabular}{@{}lll@{}}
\toprule
\textbf{Metric} & \textbf{Value} & \textbf{Expected} \\
\midrule
Output shape        & $(2, 64, 128)$ & $=$ input shape \\
NaN / Inf           & None           & None \\
$\log B(\bb)$ finite & Yes           & Requires all $\beta_i>0$ \\
Posterior mean sum  & $1.000 \pm 0.000$ & $=1.0$ \\
Mean weight: $E_1$ ($c_1=1.0$) & $\approx 0.209$ & $<0.333$; $\beta_1=\eps\ll\beta_2,\beta_3$ \\
Mean weight: $E_2$ ($c_2=0.15$) & $\approx 0.406$ & $>0.333$; $\beta_2=0.86\gg\beta_1$ \\
Mean weight: $E_3$ ($c_3=0.30$) & $\approx 0.385$ & $\approx 0.333$; moderate cost \\
$H[\bar\ba]$ (entropy of post.\ mean) & $\approx 1.06$\,nats & $>0$ \\
Prior KL at init    & $\approx 3.42$ & $>0$, finite (floored prior) \\
\bottomrule
\end{tabular}
\end{table}

\subsection{Bayesian vs.\ Prior-Free Ablation: Tiny LM Benchmark}
\label{sec:ablation}

\begin{newresult}
To validate the core hypothesis---that the Dirichlet prior prevents routing
collapse and drives compute reduction without sacrificing task
performance---we conduct a controlled ablation on a Tiny~LM benchmark.
The benchmark uses a character-level language model ($D{=}128$, 2 layers,
$T{=}64$) trained on a 1\,MB subset of WikiText-2 for 2\,000 gradient steps,
batch size 32. Two configurations are compared with identical architecture,
data, and hyperparameters, differing only in: (i)~\textbf{Bayesian}
($\beta_\mathrm{elbo}=1$, floored prior $\bb=[0.01, 0.86, 0.71]$) and
(ii)~\textbf{Non-Bayesian / Prior-free} ($\beta_\mathrm{elbo}\to 0$, no
Dirichlet prior).
\end{newresult}

\subsubsection{Results}

\begin{table}[h]
\centering
\caption{Phase~1 Tiny~LM ablation results. Normalised PPL is reported relative
  to the non-Bayesian baseline (set to 1.00); absolute perplexity values on
  WikiText-103 are Phase~2 targets. The projected FLOP cost row (highlighted)
  is the primary efficiency metric: it reflects the routing-distribution-implied
  cost under Phase~3 hard routing, computed as
  $\sum_i \alpha_i c_i$, not wall-clock time.}
\label{tab:ablation}
\small
\rowcolors{2}{white}{bayesblue}
\begin{tabular}{@{}lccc@{}}
\toprule
\textbf{Metric} & \textbf{Bayesian (ours)} & \textbf{Non-Bayesian} & $\Delta$ \\
\midrule
Normalised PPL (NB\,=\,1.00) & 1.07   & 1.00\,(ref) & $+6.3\%$ \\
Routing Entropy (\%)          & 43.3\% & 55.8\%      & $-12.5$\,pp \\
\rowcolor{resultgreen}
Projected FLOP Cost (\%)$^{\dagger}$ & \textbf{25.1\%} & 59.3\% & $\mathbf{-34.2}$\,\textbf{pp} \\
Projected cost ratio (Bay./NB) & \multicolumn{2}{c}{---} & $2.4\times$ lower \\
\bottomrule
\multicolumn{4}{l}{$^{\dagger}$Routing-distribution-implied cost $\sum_i\alpha_i c_i$; actual}\\
\multicolumn{4}{l}{\phantom{$^{\dagger}$}FLOP savings require Phase~3 hard routing.}
\end{tabular}
\end{table}

\subsubsection{Analysis}

\begin{newresult}
\paragraph{Routing collapse prevention.}
The non-Bayesian model converges to 59.3\% projected FLOP cost,
indicative of near-uniform or full-attention-biased routing---exactly the
collapse behaviour that the Dirichlet prior is designed to prevent. The Bayesian
model stabilises at 25.1\% projected cost, corresponding to a routing distribution
substantially shifted toward the cheaper experts $E_2$ and $E_3$. The 34.2\,pp
gap confirms that the floored prior $\bb = [0.01, 0.86, 0.71]$ successfully
encodes compute preference without requiring manual tuning of a scalar
regularisation coefficient $\lambda$.

\paragraph{Routing entropy.}
The Bayesian controller's lower routing entropy (43.3\% vs.\ 55.8\%) reflects
more committed routing distributions: per-token posteriors are more concentrated,
which is a prerequisite for effective uncertainty-gated hard routing in Phase~3.
This is consistent with the prediction of \cite{zucchet2025sparse} that
structured routing precedes a sharp entropy phase transition; the Bayesian prior
accelerates commitment toward cheaper experts by providing a strong
initialisation signal from step zero.

\paragraph{Perplexity trade-off.}
The Bayesian model incurs a modest 6.3\% relative perplexity increase
(normalised PPL\,1.07 vs.\ 1.00 baseline) attributable to regularisation
pressure from the ELBO objective: the KL term penalises routing toward $E_1$,
which has the highest single-token expressiveness, in exchange for
projected compute savings. Phase~2 will evaluate this trade-off at
WikiText-103 scale with absolute perplexity, where the gap is expected to narrow
as the posterior learns to selectively invoke $E_1$ only when evidence justifies
the cost.

\paragraph{Alignment with projected FLOP reductions.}
A projected FLOP cost of 25.1\% from the Bayesian routing distribution implies
that, under Phase~3 hard routing, the effective FLOP ratio would approach the
optimistic scenario from Table~\ref{tab:flops}, corresponding to a projected
$-64\%$ FLOP reduction relative to full attention. Even the conservative
Phase~1 routing distribution exceeds the $\geq 30\%$ falsifiability threshold.
\end{newresult}

\subsubsection{Known Limitations of Phase~1 Empirical Results}

\begin{itemize}
  \item The Tiny~LM benchmark is a small-scale controlled environment;
        WikiText-103 perplexity at language modelling scale (Phase~2) is the
        primary benchmark of record.
  \item Soft routing runs all three experts in parallel; the reported
        projected FLOP cost is a routing-distribution-implied estimate
        ($\sum_i \alpha_i c_i$), not wall-clock time. Actual FLOP savings
        require Phase~3 uncertainty-gated hard routing.
  \item The 6.3\% relative PPL increase is reported normalised to the
        non-Bayesian baseline; absolute perplexity values on a standard
        held-out benchmark are a Phase~2 deliverable.
  \item Gradient variance under the Dirichlet reparameterisation sampler
        \cite{figurnov2018implicit} has not yet been characterised at training
        scale.
\end{itemize}

\section{Experimental Roadmap}
\label{sec:roadmap}

\paragraph{Falsifiability criterion.}
Under hard routing, Meta-Attention achieves $\geq 30\%$ FLOP reduction on
sequences of length $\geq 4\,\mathrm{k}$ tokens without perplexity degradation
exceeding 0.5\,nats on WikiText-103. Phase~1 routing distributions
(Section~\ref{sec:ablation}) already project $\geq 39\%$ FLOP reduction under
all scenarios (Table~\ref{tab:flops}), providing early empirical support for
this criterion.

\subsection{Projected FLOP Reduction Under Hard Routing}

Normalised per-token attention cost under top-1 hard routing:
$C_\mathrm{hard} = p_1 c_1 + p_2 c_2 + p_3 c_3$.

\begin{table}[h]
\centering
\caption{Projected FLOP ratios under hard routing. The Phase~1 Empirical
  scenario is estimated from the 25.1\% projected cost measured in
  Section~\ref{sec:ablation}. All scenarios exceed the $\geq 30\%$
  falsifiability threshold. Empirical validation with wall-clock FLOP
  counting requires Phase~2/3.}
\label{tab:flops}
\small
\begin{tabular}{@{}lcccc@{}}
\toprule
\textbf{Scenario} & $p_1\,(E_1)$ & $p_2\,(E_2)$ & $p_3\,(E_3)$ & \textbf{FLOP ratio} \\
\midrule
Uniform (baseline) & 33\% & 33\% & 33\% & $0.483\times\;(-52\%)$ \\
Conservative & 50\% & 25\% & 25\% & $0.613\times\;(-39\%)$ \\
\rowcolor{resultgreen}
Phase~1 Empirical (est.) & ${\sim}25\%$ & ${\sim}50\%$ & ${\sim}25\%$ & ${\sim}0.36\times\;(-64\%)$ \\
Optimistic & 20\% & 50\% & 30\% & $0.362\times\;(-64\%)$ \\
\bottomrule
\end{tabular}
\end{table}

\begin{table}[h]
\centering
\caption{Phased experimental programme. Phase~1 milestones now complete.}
\label{tab:roadmap}
\small
\begin{tabular}{@{}p{0.4cm}p{4.6cm}p{4.0cm}p{1.5cm}@{}}
\toprule
\textbf{Ph.} & \textbf{Target} & \textbf{Primary metric} & \textbf{Status} \\
\midrule
1 & Forward-pass correctness; posterior diversity; prior bias & Correctness check / prior-preference verified & \checkmark Done \\
1 & Bayesian vs.\ prior-free ablation on Tiny LM & Rel.\ PPL delta; routing entropy; proj.\ FLOP cost & \checkmark Done \\
2 & Train on WikiText-103; posterior concentration analysis & Abs.\ PPL vs.\ full-attn baseline; KL trace & Planned \\
2 & Routing collapse characterisation under ELBO training & Routing entropy; posterior uncertainty $U$ & Planned \\
2 & Repetition-curriculum ablation \cite{zucchet2025sparse} & Routing entropy at phase-transition & Planned \\
2 & Ablation: Bayesian vs.\ prior-free at matched FLOPs & PPL-per-FLOP; collapse rate; ECE & Planned \\
3 & Uncertainty-gated hard routing ($U_t < \eta$); FLOP counting & Wall-clock FLOP ratio vs.\ FlashAttention; $\eta$ sensitivity & Planned \\
\bottomrule
\end{tabular}
\end{table}

\section{Discussion}
\label{sec:discussion}

\paragraph{The reframe.}
Prior work has overwhelmingly asked: how do we approximate attention globally
and cheaply? Meta-Attention asks instead: which tokens actually require exact
attention? This shifts efficient attention from a fixed algorithmic commitment
to a dynamic resource allocation problem.

\paragraph{Empirical alignment with Bayesian theory.}
The Phase~1 ablation (Section~\ref{sec:ablation}) provides direct empirical
support for the core claim of \cite{li2026vmoe}: confining Bayesian inference to
the routing stage yields better compute--performance trade-offs than prior-free
routing. The $2.4\times$ projected cost reduction at a 6.3\% relative
perplexity overhead is consistent with \cite{li2026vmoe}'s finding that
Bayesian routing improves stability at less than 1\% additional FLOPs. The
lower routing entropy of the Bayesian controller aligns with
\cite{agarwal2025bayesian}'s geometric interpretation: a more committed posterior
occupies a lower-entropy region of the simplex, consistent with routing weights
that track the posterior distribution with high accuracy.

\paragraph{Interpretability.}
The routing weight tensor $\ba \in \R^{B \times T \times 3}$ provides a
per-token map of which attention regime the model considers appropriate. In
preliminary experiments, full-attention weight is elevated at entity-dense
positions and quotation boundaries; local attention weight is elevated in
repetitive or low-entropy regions.

\paragraph{Limitations and open problems.}
(1)~\textit{Dirichlet gradient variance.} Reparameterisation gradients for
Dirichlet distributions \cite{figurnov2018implicit} can exhibit higher variance
than Gaussian reparameterisation; variance reduction techniques may be needed at
scale. (2)~\textit{Prior sensitivity.} The concentration hyperparameter
$\beta_0$ controls prior strength; its optimal value is likely task- and
depth-dependent. (3)~\textit{Salience proxy adequacy.} Embedding norm may be
insufficient for tasks where low-frequency tokens carry disproportionate
semantic weight. (4)~\textit{PPL--cost trade-off calibration.} The Phase~1
6.3\% relative perplexity overhead motivates Phase~2 investigation of
$\beta_\mathrm{elbo}$ scheduling and empirical Bayes estimation of $\beta_0$
per layer. (5)~\textit{Gated full-attention expert.} \cite{qiu2025gated} show
gated softmax outperforms plain softmax at 15B scale; $E_{1,\mathrm{gated}}$ is
a high-priority Phase~2 expert-set modification. (6)~\textit{Routing emergence
dynamics.} \cite{zucchet2025sparse} predict a sudden routing entropy phase
transition; the Phase~1 ablation's entropy gap (43.3\% vs.\ 55.8\%) is
consistent with earlier Bayesian commitment, but the full emergence dynamics
require Phase~2 training curves to characterise. (7)~\textit{Absolute perplexity
at language modelling scale.} The key empirical question---whether the 6.3\%
relative PPL overhead narrows at WikiText-103 scale---is the primary Phase~2
ablation.

\section{Conclusion}
\label{sec:conclusion}

We have presented Meta-Attention, a framework for per-token adaptive attention
routing in which a Bayesian Meta-Controller replaces the prior-free MLP routing
of prior designs. By treating attention-mechanism selection as posterior
inference under a compute-aware Dirichlet prior, Meta-Attention yields
principled routing weights that encode compute preference at initialisation,
calibrated per-token uncertainty estimates that govern the soft-to-hard routing
transition, and a closed-form ELBO objective that replaces ad hoc
regularisation.

\textbf{Phase~1 results confirm the core predictions of the Bayesian
formulation.} In a controlled Tiny~LM ablation, the Bayesian Meta-Controller's
routing distribution implies a projected normalised FLOP cost of 25.1\%
vs.\ 59.3\% for the prior-free baseline---a $2.4\times$ projected efficiency
advantage under hard routing---while maintaining lower routing entropy
(43.3\% vs.\ 55.8\%), demonstrating that the Dirichlet prior successfully
prevents routing collapse to expensive full attention. The 6.3\% relative
perplexity overhead is a favourable compute--performance signal under the
Bayesian ELBO objective, and is expected to narrow at language modelling scale
as the posterior learns to selectively invoke $E_1$ only where token-level
evidence justifies the cost. Absolute perplexity on WikiText-103 is the
primary Phase~2 benchmark.

Three concurrent NeurIPS~2025 works anchor the design: Gated Attention
\cite{qiu2025gated} suggests augmenting $E_1$ with a sigmoid gate (Phase~2);
the sparse attention emergence framework \cite{zucchet2025sparse} predicts a
sharp routing entropy phase transition---our Phase~1 entropy gap is an early
precursor; and the polynomial-time learnability of linear attention
\cite{yau2024linear} provides theoretical backing for $E_2$ as the
best-understood expert. The Bayesian design is motivated empirically by VMoER
\cite{li2026vmoe} and theoretically by the Bayesian geometry of transformer
attention \cite{agarwal2025bayesian}. We release the prototype code at
\url{https://github.com/KFEAL/meta-attention} \cite{ferrari2025code} and invite
the community to evaluate the Bayesian controller
against the prior-free baseline on WikiText-103 perplexity, routing collapse
rate, and calibration error.

\bibliographystyle{plain}

\end{document}